\documentclass{article}
\usepackage[dblblindworkshop,final]{neurips_2026}
\workshoptitle{Agentic AI for Biological Discovery}
\usepackage[T1]{fontenc}
\usepackage[utf8]{inputenc}
\usepackage{amsmath,amssymb,booktabs,graphicx}
\usepackage{algorithm,algpseudocode}
\usepackage{url}
\usepackage[hidelinks]{hyperref}

\makeatletter
\renewcommand{\@notice}{}
\makeatother
\title{Historical-Feature Residual Adaptation\\ for Sequential Screening}
\author{%
  Yixiang Hong \quad Yuchen Zhang \\
  The University of Texas Health Science Center at Houston \\
  Houston, Texas, USA
}

\hypersetup{
  pdftitle={Historical-Feature Residual Adaptation for Sequential Screening},
  pdfauthor={Yixiang Hong and Yuchen Zhang},
  pdfsubject={NeurIPS 2026 AgenticLS Workshop, Paper 87}
}
\begin{document}
\maketitle

\begin{abstract}
Sequential screening seeks to discover as many positive outcomes as possible under a fixed experiment budget. We study acquisition-time residual adaptation: preserve a reusable base scorer and learn a ranking correction in fixed historical features, using the base scores recorded before each batch's feedback as fitting anchors. Each decision combines the current base scores with the correction learned from earlier observations. In the primary evaluation on 20 test screens from 15 sources, this procedure increases source-macro hit recovery from 22.58\% to 28.61\%, a 26.7\% relative improvement, and exceeds same-feature pairwise fitting by 2.88 percentage points. It also outperforms static-anchor and tuned-logistic controls. Across the recorded budget checkpoints from 500 to 1,000 requests, it leads the released scorer, pairwise fitting and tuned logistic. A supporting study with independent features finds matched-base gains for random-forest and gradient-boosting scorers on 21 tasks from six held-out sources. The results support preserving a reusable scorer while adapting its ranking through acquisition-time residuals.
\end{abstract}

\section{Introduction}
A sequential screening system must decide which experiments to run next using the outcomes it has already acquired. Under a limited budget, the objective is to recover useful discoveries, so the quality of the evolving acquisition policy matters directly. Historical screens can provide both an initial ranking and a representation for adapting that ranking as observations accumulate. The central question is how to combine this reusable information with feedback from the current task.

A base scorer already embodies useful structure. Replacing it with a newly fitted head can discard that structure, while leaving its adaptation mechanism unchanged may fail to exploit patterns present in historical data. Residual adaptation offers a middle ground: retain the base scores and learn a small correction. In a sequential setting, however, the base itself can change after each observation. A score recomputed after receiving feedback is not the prediction that preceded that feedback. The choice of which prediction to use as a fitting offset therefore becomes part of the adaptation rule.

\begin{figure}[t]
\centering
\includegraphics[width=\linewidth]{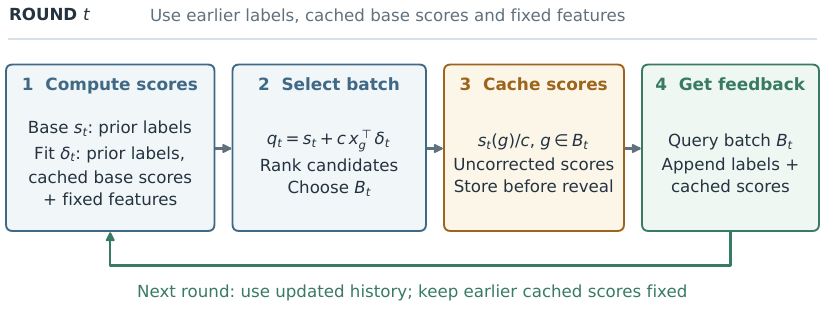}
\caption{Acquisition-time residual adaptation. The current base score and a correction fitted from earlier feedback determine the next batch. Its uncorrected base scores are stored before feedback is revealed. These immutable anchors and the selected outcomes enter the history for subsequent rounds. Historical features remain fixed.}
\label{fig:overview}
\end{figure}

We study \emph{acquisition-time residual adaptation}. Before a selected batch is observed, the policy records its uncorrected base scores. It then fits a pairwise ranking correction against these recorded scores using fixed historical candidate features. Each subsequent decision combines the correction with the base's current scores. This separates a reusable scoring interface from the model that supplies the scores and from the representation used by the correction. Figure~\ref{fig:overview} illustrates the resulting closed loop.

Our primary instantiation uses AssayFormer, a released feedback-conditioned scorer for AssayBench-Loop \citep{assaybench,assayloop}. We augment its ten-dimensional coordinates with 64 historical directions and update only the residual head, retaining the pretrained parameters. Matched comparisons test whether this procedure adds value beyond ordinary fitting, fixed score anchors, and simpler score combinations. This is the primary evaluation of the complete preserve-base residual design. A supporting portability study uses independently constructed features and previously fixed random-forest and gradient-boosting configurations on 21 tasks from six sources excluded from feature training and earlier evaluations. Supplementary comparisons and earlier development results appear in the supplement.

The paper makes three contributions:
\begin{itemize}
\item A preserve-base residual interface that fits corrections against predictions recorded before feedback, using fixed historical features and no access to the base model's internal parameters during online residual updates.
\item A primary matched-budget evaluation showing a 6.03-point gain over the released policy, a 2.88-point gain over same-feature pairwise fitting, and a 3.60-point gain over static-anchor adaptation, together with a retained advantage over tuned logistic regression.
\item Supporting portability evidence from two independently trained tree ensembles on held-out sources.
\end{itemize}
These contributions concern the experiment-selection component of a closed-loop discovery system: observed feedback updates the policy that proposes the next batch. The evaluation isolates this component through retrospective replay, making its information inputs and request costs explicit.

\section{Related work}
\paragraph{Biomedical prediction and sequential discovery.}
Biomedical prediction has used tree ensembles on population health data \citep{liu2025nhanes} and personalized federated models for clinical risk prediction \citep{zhao2026demography}. In sequential discovery, active search evaluates a policy by the number of positives it discovers under a query budget; batch active search also accounts for decisions made before a group of outcomes becomes available \citep{jiang2018}. Predictive active learning instead seeks informative labels for improving a model. Learning Active Learning learns a selection strategy from previous learning outcomes \citep{konyushkova2017}. Our objective follows budgeted discovery. We adapt the current ranking from acquired feedback rather than optimize a predictive-error criterion or introduce a look-ahead acquisition rule.

\paragraph{Historical representations and adaptation.}
AssayLoop learns acquisition policies across historical screens and includes matrix-factorization representations and Bayesian factor-model baselines \citep{assayloop}. REAP combines frozen representations with task-specific ranking and sequential measurements \citep{reap}. These approaches motivate separating reusable representations from adaptation on a new task. Our independent-feature protocol further separates that representation from a particular pretrained policy.

\paragraph{Residual learning and ranking.}
Residual connections are established components of convolutional networks \citep{he2016resnet} and Transformers \citep{vaswani2017attention}. Attention-based architectures have been applied to medical image segmentation \citep{hatamizadeh2022unetr}, protein structure prediction \citep{jumper2021alphafold}, and speech emotion recognition \citep{zhao2026residualattention}. Additive model adaptation \citep{gao2009} and pairwise ranking losses \citep{ranknet} provide established building blocks. TTT-Embed optimizes residual vectors outside a frozen retrieval encoder using ranking rewards \citep{tttembed}. Our contribution is the study of historical features and acquisition-time score anchors within a sequential screening loop. Same-feature pairwise, static-anchor, logistic and fusion controls distinguish this procedure from simply fitting another ranking head.

\section{Historical-feature residual adaptation}
\subsection{A score interface for the discovery loop}
Let \(U_t\) be the unqueried candidates before batch \(t\), and let \(\mathcal D_{<t}\) contain the policy's previously acquired observations. A base scorer supplies \(s_t(g)\) for \(g\in U_t\), using only this history. A pretrained base may retain its parameters while conditioning on feedback; an online base may be refitted on \(\mathcal D_{<t}\). The adapter also receives a fixed feature vector \(x_g\in\mathbb R^d\) and a positive reference scale \(c\), both constructed before target outcomes are revealed.

Before revealing the feedback for a selected batch \(B_b\), cache \(a_{bg}=s_b(g)/c\) for its candidates. Within each observed batch, pair a positive outcome \(i\) with a negative outcome \(j\); collect these pairs in \(\mathcal P_t\). The correction solves
\begin{align}
 \delta_t&=\arg\min_\delta\ \frac{1}{|\mathcal P_t|}
 \sum_{(b,i,j)\in\mathcal P_t}
 \log\!\left(1+\exp\!\left[-(a_{bi}-a_{bj}+\delta^\top(x_i-x_j))\right]\right)
 +\tfrac12\|\delta\|_2^2,\label{eq:loss}\\
 q_t(g)&=s_t(g)+c\,\delta_t^\top x_g.\label{eq:score}
\end{align}
The next batch consists of the highest-scoring unqueried candidates under \(q_t\), with the common initialization described below and the seeded tie-breaking rule in Appendix~\ref{app:implementation}. Each pair uses predictions made under the same acquisition history. The cached offsets exclude the previous correction and are never overwritten in the primary method. Fitting starts from zero with fixed regularization; no eligible pairs or a rejected fit gives a zero correction and retains the base's prescribed fallback.

\begin{algorithm}[b]
\caption{Acquisition-time residual adaptation}
\label{alg:online_procedure}
\begin{algorithmic}[1]
\Require Candidate pool \(U\), fixed features \(\{x_g\}\), scale \(c>0\), base scorer,
shared first batch \(B_{\mathrm{init}}\), batch size \(k\), rounds \(T\) (budget \(kT\))
\State \(\mathcal D\gets\varnothing;\quad \mathcal A\gets\varnothing\)
\For{\(t=1,\ldots,T\)}
    \State \(s_t\gets\operatorname{BaseScore}(U;\mathcal D)\)
    \State \(\delta_t\gets\operatorname{FitResidual}(\mathcal D,\mathcal A;\{x_g\})\)
    \State \(q_t(g)\gets s_t(g)+c\,\delta_t^\top x_g\), for \(g\in U\)
    \State \(B_t\gets B_{\mathrm{init}}\) if \(t=1\), otherwise \(\operatorname{Select}_k(U,q_t)\)
    \State \(\mathcal A\gets\mathcal A\cup\{(t,g,s_t(g)/c):g\in B_t\}\) \Comment{Before feedback}
    \State \(\{y_g:g\in B_t\}\gets\operatorname{Query}(B_t)\)
    \State \(\mathcal D\gets\mathcal D\cup\{(t,g,y_g):g\in B_t\};\quad U\gets U\setminus B_t\)
\EndFor
\end{algorithmic}
\smallskip
\textit{FitResidual} solves Equation~\ref{eq:loss} from zero using earlier within-batch pairs and cached offsets; it returns zero if no eligible pairs exist or the fit is rejected. \textit{Select} chooses the \(k\) highest-ranked unqueried candidates, with seeded ties and the prescribed base fallback. Earlier cached offsets are never overwritten.
\end{algorithm}

Algorithm~\ref{alg:online_procedure} summarizes the update order: cache uncorrected scores before querying feedback and retain them for subsequent fits.

\paragraph{What the anchor changes.}
Ordinary pairwise fitting sets the offsets to zero and learns the entire ranking from observed labels. A static-anchor residual uses the base's empty-history predictions both for fitting offsets and for subsequent candidate scoring. Acquisition-time anchors instead retain how the evolving base ranked each batch before observing it, so the residual is fitted conditionally on that prediction. This construction does not require calibrated probabilities. Its usefulness is an empirical question because deployment combines the learned correction with the current base, which may differ from earlier bases.

\subsection{Fixed coordinates from historical screens}
Let \(M\) and \(H\) be candidate-by-historical-screen measurement and hit matrices, with both entries zero for an unmeasured candidate. If \(p_0\) is the pooled measured hit rate, define
\begin{equation}
 p_g=\frac{\sum_s H_{gs}+20p_0}{\sum_s M_{gs}+20},
 \qquad T_{gs}=H_{gs}-M_{gs}p_g.
\end{equation}
Column-center \(T\) and \(M\) to obtain \(T_c\) and \(C\). Equal Frobenius normalization gives
\begin{equation}
 X=\frac{1}{\sqrt{2}}\left[T_c/\|T_c\|_F,\ C/\|C\|_F\right].
\end{equation}
A fixed low-rank decomposition of this table supplies candidate coordinates. The two instantiations below use the same residual objective but construct their coordinates independently.

\paragraph{Pretrained-policy instantiation.}
The primary study uses 1,349 training screens across 134 sources and a fixed universe of 20,893 candidates. The released AssayFormer coordinates \(E\in\mathbb R^{n\times10}\) give scores of the form \(\mathbf s_t=E\mathbf u_t+\mathbf b\), where \(\mathbf u_t\) depends on acquired feedback and \(\mathbf b\) is fixed. Consequently,
\begin{equation}
 \mathbf s_t-\mathbf s_0=E(\mathbf u_t-\mathbf u_0)\in\operatorname{col}(E).
 \label{eq:span}
\end{equation}
Centering and scaling gives \(V=(E-\mathbf1\mu^\top)D^{-1}\), with \(\operatorname{col}([\mathbf1,V])=\operatorname{col}([\mathbf1,E])\). We project \(X\) off this space, retain 64 singular directions, standardize them and concatenate them with \(V\), yielding 74D features. The added directions enlarge the space available to the correction; performance is tested against a same-dimensional unprojected control. Because the original ten coordinates are retained, the full residual is not constrained to be orthogonal to the base scores.

The scale \(c\) is the population standard deviation of empty-history base scores, floored at \(10^{-8}\). Pretrained parameters, feature transformation and scale remain fixed. The base continues to condition on feedback while the 74-parameter residual is updated online. A mask-only control decomposes \(C\); the unprojected control omits the projection while retaining ten original and 64 historical coordinates.

\paragraph{Independent-feature instantiation.}
A separate protocol constructs \(X\) from restricted historical data, applies rank-64 SVD directly, standardizes the coordinates and appends one standardized historical-prior logit, giving 65D features. No AssayFormer coordinates, predictions, vocabulary, description embeddings or initial batches enter this construction. Random-forest and histogram gradient-boosting (HGB) bases are refitted on their own acquired observations using the same features; probabilities are converted to clipped logits. The residual scale is fixed to the historical-prior-logit standard deviation. Appendix~\ref{app:independent} gives the base settings; Appendix~\ref{app:heldout} reports the supporting held-out-source comparisons.

\section{Experimental design}
\paragraph{Feedback and budget.}
Every trajectory requests ten batches of 100 distinct candidates. Methods share their initial batch within each study and seed, then receive only their own acquired feedback. The evaluator retains target-library membership and unrevealed labels. An unmeasured request returns negative feedback and consumes a slot. Table~\ref{tab:study_design} separates the two studies' representations, initialization and hit denominators.

\begin{table}[htbp]
\centering
\caption{Primary evaluation and supporting portability study. The latter uses previously fixed methods on six previously unused sources within the same benchmark. Absolute FH values are compared within studies because cohorts and denominators differ.}
\label{tab:study_design}
\small
\begin{tabular}{p{.20\linewidth}p{.35\linewidth}p{.35\linewidth}}
\toprule
& Primary pretrained-policy study & Supporting portability study\\
\midrule
Historical data & 1,349 screens / 134 sources & 852 screens / 29 sources\\
Representation & Former + complementary history; 74D & History decomposition + prior; 65D\\
Evaluation & 20 test screens / 15 sources & 21 held-out tasks / 6 sources\\
Base updates & Fixed parameters; feedback-conditioned scores & Numerical bases refitted on acquired labels\\
Initial batch & Released-policy batch & Uniform random, shared within seed\\
Acquisition seeds & One & Three\\
FH denominator & Full reference hit library & Positives reachable in training-derived universe\\
\bottomrule
\end{tabular}
\end{table}

\paragraph{Metrics and aggregation.}
The original study reports fraction of full-library hits recovered (FH), inherited enrichment factor (EF), and requested-budget EF. FH includes reference hits outside its fixed universe. Inherited EF uses the effective measured-selection budget, whereas requested-budget EF charges all 1,000 requests; Appendix~\ref{app:implementation} specifies both metrics and zero-denominator handling. The independent study uses reachable-hit FH under its training-derived universe, with coverage reported separately. We average seeds within screen when applicable, screens within recorded source, and sources equally. Seed differences and single-source omissions are descriptive robustness analyses; acquisition seeds are not additional biological sources.

\paragraph{Pretrained-policy comparison.}
The original methods and settings were fixed before accessing outcomes from the 20 official-test screens. Table~\ref{tab:main} focuses on the released scorer, same-feature pairwise fitting, a static-anchor residual, and the proposed method. A tuned logistic comparator was subsequently selected on development data and evaluated on the exposed test cohort. Paired source-bootstrap intervals are marginal descriptive 95\% intervals, without multiplicity adjustment. The common local universe differs from the source paper's universe, so all reported comparisons use matched replay. Appendix~\ref{app:provenance} summarizes chronology and additional controls.

\paragraph{Supporting held-out-source portability.}
We retain the 65D features and 21,069-candidate universe constructed from 852 historical screens across 29 sources. The confirmation cohort contains 21 cancer drug/chemical-response knockout tasks from six recorded sources excluded from feature training and all earlier evaluation cohorts. Tasks were selected using design metadata, with all selected tasks retained. The base configurations, residual objective, feature dimension, anchor rule, hyperparameters and endpoint were fixed before accessing these outcomes. We report five policies: matched RF/HGB pairs and ordinary pairwise fitting, evaluated across three acquisition seeds for 315 trajectories. Table~\ref{tab:heldout} presents the tree ensembles; Appendix~\ref{app:heldout} includes the pairwise control and source-separation details. Earlier experiments on 13 reused development screens provide supporting development evidence (Appendix~\ref{app:independent}).

\section{Results}
The primary evaluation tests whether preserving the pretrained scorer and using acquisition-time anchors improves discovery beyond replacement fitting and static anchors. A separate held-out-source study provides supporting evidence for portability to other scorers.

\subsection{Primary evaluation of the pretrained screening policy}
\begin{table}[htbp]
\centering
\caption{Key pretrained-policy comparisons on 20 test screens from 15 sources after 1,000 requests. Metrics are source-macro averages; the last column gives the proposed method's FH advantage. The final row is a later, development-selected comparator evaluated on the exposed cohort. Additional controls appear in Appendix~\ref{app:provenance}.}
\label{tab:main}
\small
\begin{tabular}{lrrrr}
\toprule
Method & FH (\%) & EF & Budget EF & $\Delta$FH (pp)\\
\midrule
Released AssayFormer & 22.5837 & 4.5655 & 4.3026 & +6.03\\
74D ordinary pairwise & 25.7349 & 5.0797 & 4.9382 & +2.88\\
74D static-anchor residual & 25.0171 & 4.9515 & 4.7699 & +3.60\\
\textbf{74D historical-feature residual} & \textbf{28.6144} & \textbf{5.7079} & \textbf{5.4880} & ---\\
\midrule
74D tuned logistic (supplementary) & 26.4220 & 5.2525 & 5.0699 & +2.19\\
\bottomrule
\end{tabular}
\end{table}

\paragraph{More discoveries at a fixed budget.}
Historical-feature residual adaptation improves FH from 22.58\% to 28.61\% relative to the released policy: +6.03 percentage points, or 26.7\% relative improvement (Table~\ref{tab:main}). The descriptive paired-source interval is [2.46, 11.19] points. The method wins on 13 sources, ties on one and loses on one. Requested-budget EF rises from 4.30 to 5.49, showing that the gain persists when all requested selections are charged.

\paragraph{Beyond ordinary fitting and fixed anchors.}
The same-feature pairwise head reaches 25.73\% FH and the static-anchor residual reaches 25.02\%. Acquisition-time residual adaptation improves on these by 2.88 points [1.20, 4.76] and 3.60 points [0.72, 7.84], respectively. Tuned logistic regression reaches 26.42\% FH; the residual retains a 2.19-point advantage with ten source wins, two ties and three losses. Removing any one source leaves +1.76 to +2.55 points. These comparisons support preserving the base scorer and adapting its ranking through feedback-conditioned residuals.

\begin{figure}[t]
\centering
\includegraphics[width=\linewidth]{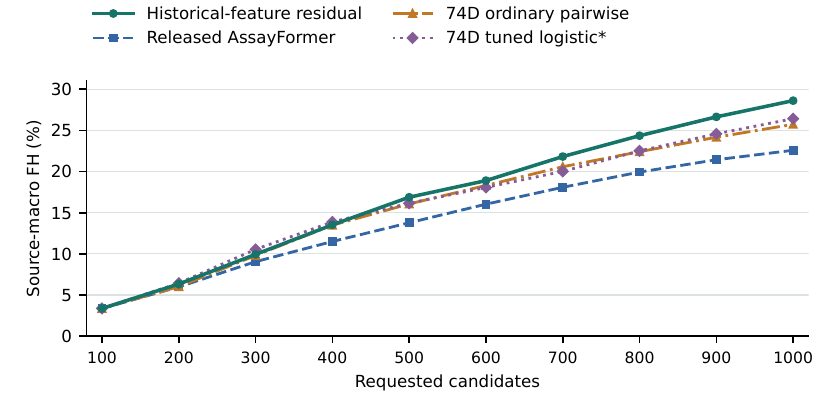}
\caption{Hit recovery versus requested budget on the same 20 test screens and 15 sources. Each curve averages screens within source and then sources equally, using full reference-hit denominators. All ten recorded budget checkpoints are shown. Tuned logistic is the later development-selected comparator.}
\label{fig:budget}
\end{figure}

\paragraph{Recovery across the acquisition budget.}
The residual leads all three plotted controls at every recorded checkpoint from 500 through 1,000 requests (Figure~\ref{fig:budget}), extending the endpoint advantage across the latter half of the budget. The curves retain all ten checkpoints: policies share the first batch, and tuned logistic leads the residual at 200--400 requests.

\paragraph{Additional representation controls.}
Rank fusion reaches 26.02\% FH and the unprojected residual reaches 26.78\% (Appendix~\ref{app:provenance}). The projection advantage is concentrated in one source, so these results support the complete adaptation recipe without establishing projection alone as a robust mechanism across sources.

\subsection{Supporting portability on held-out sources}
\begin{table}[H]
\centering
\caption{Held-out-source portability on 21 tasks from six previously unused recorded sources: three acquisition seeds and 1,000 requests per trajectory. We report source-macro reachable FH, SD across seed means, and gains over each matched base. The comparison including ordinary pairwise fitting appears in Table~\ref{tab:heldout_all}.}
\label{tab:heldout}
\small
\setlength{\tabcolsep}{8pt}
\begin{tabular}{lrr}
\toprule
Policy & Reachable FH (\%) $\pm$ seed SD & Gain (pp) \\
\midrule
Random forest & $17.1853 \pm 0.4700$ & --- \\
\textbf{Random forest + residual} & \textbf{19.9836} $\pm$ \textbf{1.8659} & \textbf{+2.80} \\
\addlinespace[2pt]
HGB & $17.9108 \pm 1.6121$ & --- \\
\textbf{HGB + residual} & \textbf{18.6596} $\pm$ \textbf{1.8026} & \textbf{+0.75} \\
\bottomrule
\end{tabular}
\end{table}

\paragraph{Fixed-method gains on new sources.}
With the original feature table and residual settings, random-forest FH rises from 17.19\% to 19.98\% (+2.80 points; Table~\ref{tab:heldout}). All six sources improve, as do all three acquisition seeds (+4.48, +1.98 and +1.94 points). Removing any one source retains +1.65 to +3.32 points. This reproduces the own-base adaptation gain on sources excluded from the historical representation and earlier method development.

HGB recovery rises from 17.91\% to 18.66\% (+0.75 points). All three seed averages improve (+0.86, +0.97 and +0.41), while the six source effects split evenly between gains and declines (Appendix~\ref{app:heldout}, Figure~\ref{fig:heldout_sources}). Both comparisons reuse the same residual objective and acquisition-time anchor rule without adjustment on the confirmation cohort.

\paragraph{What the comparison establishes.}
The confirmation supports gains over the matched RF/HGB bases within the same benchmark. Ordinary pairwise fitting reaches 20.17\% FH, so these improvements do not establish superiority over every same-feature policy. Appendix~\ref{app:heldout} reports the five-policy comparison and descriptive source intervals. The earlier 13-screen results are reported separately as development evidence.

\section{Discussion and conclusion}
The primary finding is improved discovery from preserving the base scorer and fitting residuals against acquisition-time predictions. At the same 1,000-request budget, source-macro hit recovery rises by 6.03 points over the released policy, with advantages over same-feature pairwise fitting, static anchors and tuned logistic. The residual leads all three plotted controls at recorded checkpoints from 500 through 1,000 requests. These matched comparisons support the complete adaptation design.

The held-out-source study provides supporting portability evidence within the same benchmark. Previously fixed configurations improve average recovery over matched RF/HGB bases, with particularly consistent source effects for RF. These findings extend the scope of the primary result while keeping the main empirical claim centered on the pretrained-policy comparison.

The interface fits the experiment-selection component of a discovery system: a planner supplies the task and budget, the scorer and adapter select experiments, and returned feedback updates the next decision. Retrospective replay isolates this numerical loop and provides a basis for prospective evaluation within tool-using discovery agents.

\clearpage
\bibliographystyle{plainnat}
\begingroup
\raggedright
\bibliography{references}
\endgroup
\clearpage
\appendix
\section{Implementation and comparator details}\label{app:implementation}
\paragraph{Fixed historical representation.}
The original ten coordinates retain the released non-unknown vocabulary's mean and population standard deviation. A QR basis for the intercept and those coordinates defines the projection. ARPACK retains 64 singular components with tolerance \(10^{-8}\), maximum 10,000 iterations, a fixed starting vector, and deterministic signs. Singular coordinates are centered and standardized before concatenation. The value-and-mask and mask-only representations are constructed once from training screens. The unprojected control uses the same history construction, dimensions, and standardization, omitting the projection step.

\paragraph{Residual solver and fallbacks.}
The objective has penalty one. L-BFGS-B starts from zero each round, with maxiter 100, gtol \(10^{-8}\), ftol \(2.220446049250313\times10^{-9}\), maxcor 10, maxfun 15,000, and maxls 20. Acceptance requires optimizer success, finite values, and an objective no worse than the zero correction plus \(10^{-12}\). No-class or no-pair rounds use the specified uncorrected policy. Pairwise terms are a fitting objective rather than independent samples or calibrated probabilities.

\paragraph{Tuned logistic and fusion comparators.}
Development evaluated seven regularization settings for ordinary pairwise and unpenalized-intercept logistic heads using leave-one-source-out parameter selection. All logistic folds selected \(\lambda=10\). The fusion development procedure selected Former weight 0.75 and logistic \(\lambda=10\) in every source fold. For the current \(N\) candidates, it combines fractional ranks \(r=(\operatorname{rank}-0.5)/N\), with average ties, as \(0.75r_{\rm Former}+0.25r_{\rm logistic}\); higher values rank first. A rejected or single-class logistic fit falls back to empty-history Former scores for the ordinary logistic policy and current Former scores for fusion. These procedures selected comparator settings, while the original residual method remained fixed. The representation and backbone were not refitted inside the source folds. Tuned logistic and fusion use maxiter 500; the supplementary unprojected arm uses the original residual's \(\lambda=1\), maxiter 100 solver. This solver match was fixed before supplementary outcomes; original test and development fits had terminated well within that cap.

\paragraph{Inputs and candidate universe.}
The 20,893-gene universe was fixed during development from candidate metadata and is not filtered by each target library. Target-library gene coverage is 96.08--100\%, while hit denominators retain the full library. The policy receives the cached original short-description embedding and its acquired Boolean feedback. The ONNX checkpoint, runtime, normalization, and candidate ordering remain fixed. The evaluation covers the numerical scorer and adaptation loop; it does not evaluate the complete LLM-to-Former system.

\paragraph{Metric definitions.}
In the pretrained-policy study, let \(L\) be the nonempty measured target library, \(H\subseteq L\) its reference hits, \(A\) the acquired candidates, and \(R\) the requested slots. Write \(h=|A\cap H|\), \(m=|A\cap L|\), and \(p=|H|/|L|\). For the complete, distinct, in-universe requests used here, FH is \(h/|H|\), inherited EF is \(h/(mp)\), and requested-budget EF is \(h/(Rp)\); each returns zero when its denominator is zero. The independent-feature study instead uses reachable-hit FH, as defined in the main text.

\paragraph{Tie-breaking.}
For each scored selection, every remaining candidate receives a fresh uniform random tie key from the acquisition-seeded generator, in fixed candidate order. Sorting uses descending acquisition score, then ascending tie key; remaining exact ties use gene identifiers in the pretrained-policy study and fixed candidate indices in the independent-feature study.

\section{Additional comparisons and evaluation scope}\label{app:provenance}
\paragraph{Evaluation chronology.}
The primary official-test comparison was fixed before target outcomes were accessed. Tuned logistic, rank fusion and unprojected residual were added later with development-selected settings and evaluated on the exposed cohort. Earlier independent-feature experiments reuse development tasks. The later source-held-out confirmation applies the original fixed methods to a separate cohort (Appendix~\ref{app:heldout}).

\paragraph{Uncertainty and verification.}
Descriptive intervals resample the 15 recorded sources with replacement 10,000 times (PCG64 seed 20260919), using common draws for the original contrasts and linear 2.5/97.5 percentiles. These are marginal intervals without multiplicity adjustment. Independent saved-output checks verified acquisition chronology, cached uncorrected offsets, cumulative metrics and source aggregation. The manuscript distinguishes the primary evaluation, supporting held-out-source study and earlier development comparisons.

\begin{table}[htbp]
\centering
\caption{Additional pretrained-policy controls on the same 20 test screens and 15 sources. Values and the primary-minus-control gap use source-macro aggregation after 1,000 requests. Fusion and the unprojected residual are later comparisons.}
\label{tab:additional_controls}
\small
\begin{tabular}{lrrrr}
\toprule
Method & FH (\%) & EF & Budget EF & $\Delta$FH (pp)\\
\midrule
74D mask-only residual & 25.4402 & 5.0179 & 4.8539 & +3.17\\
AssayFormer--logistic rank fusion & 26.0214 & 5.1187 & 4.9649 & +2.59\\
74D unprojected residual & 26.7796 & 5.2769 & 5.1093 & +1.83\\
\bottomrule
\end{tabular}
\end{table}

\paragraph{Interpreting representation and fusion effects.}
The mask-only gap has a descriptive interval of [$-0.23$, 8.06] FH points. One source accounts for 98.77\% of the projected residual's net advantage over the unprojected residual; omitting it leaves +0.024 points. That source also contributes 74.88\% of the advantage over fusion. On the reused development cohort, fusion (27.04\%) and unprojected adaptation (26.96\%) slightly exceed the primary method (26.58\%). The ranking therefore depends on the evaluation split, and these controls do not separately establish robust benefits from hit values beyond coverage or from projection alone.

\section{Independent-feature implementation and development results}\label{app:independent}
\paragraph{Historical features and coverage.}
The 852 unconditioned cell-fitness training screens span 29 recorded sources, disjoint from both independent-feature evaluation cohorts. The candidate universe is the sorted union of training libraries (21,069 entries), with no target-library filtering. Value and measurement blocks follow the main-text construction. Direct rank-64 SVD is centered and standardized, and a standardized historical-prior logit is appended. Probabilities are clipped to $[10^{-6},1-10^{-6}]$. Feature construction uses no AssayFormer assets. On the original 13-screen development cohort, this universe covers 99.8948--100\% of each target library and 99.7545--100\% of its positives. Its forest residual has 31.2516\% whole-library FH versus 31.2713\% reachable-hit FH. The same frozen feature table is reused for the 21-task confirmation, whose coverage is reported separately in Appendix~\ref{app:heldout}.

\paragraph{Matched base scorers.}
All bases are refitted from scratch on their own acquired observations. Random forest uses scikit-learn 1.7.2 with 100 trees, Gini splitting, square-root feature subsampling, bootstrap sampling, unrestricted depth, minimum split size two, minimum leaf size one, no class weighting, one worker and internal random state zero. HGB uses binary log loss, 100 iterations, learning rate 0.1, at most 31 leaves, minimum leaf size 20, at most 255 bins, zero leaf L2 penalty, all features at each split, no depth limit or class weighting, and random state zero; early stopping and warm starts are disabled. Every raw/corrected pair uses the identical base fitter. Fewer than two classes or a rejected base fit uses historical-prior scores.

\paragraph{Residuals and acquisition.}
RF and HGB probabilities are clipped to $[10^{-6},1-10^{-6}]$ before conversion to logits. The historical-prior-logit SD fixes $c$, and the uncorrected logits divided by $c$ are cached before feedback. Residuals use Equation~\ref{eq:loss} with penalty one and no base-specific adjustment. L-BFGS-B uses maxiter 500, gtol $10^{-8}$ and ftol $2.220446049250313\times10^{-9}$, with finite, converged, non-increasing-objective acceptance. Ordinary pairwise uses identical features and pair construction with zero offsets. Acquisition seeds 0, 1 and 2 share a uniform initial batch across policies; each trajectory requests ten batches of 100 distinct candidates. Subsequent selections see only that policy's acquired Boolean feedback. Requests outside a target library consume a slot and return negative feedback.

\paragraph{Earlier development results.}
Table~\ref{tab:development_portability} reports tree-ensemble adaptation and reference policies on 13 reused development screens from six sources. RF improves from 28.87\% to 31.27\%, and the later fixed-configuration HGB comparison improves from 29.55\% to 31.31\%. Both gains are positive in all three acquisition seeds. These development observations motivated a separate confirmation with fixed methods; they are not pooled with that cohort. The seven reported policies account for 273 trajectories, including 78 from the later HGB comparison.

\begin{table}[htbp]
\centering
\caption{Earlier development results for tree-ensemble adaptation and reference policies. Each policy uses 13 reused development screens, three seeds and 1,000 requests. FH is source-macro reachable-hit recovery; seed SD is descriptive.}
\label{tab:development_portability}
\small
\begin{tabular}{lrrr}
\toprule
Method & FH (\%) & Seed SD (pp) & Own-base gain (pp)\\
\midrule
Uniform random & 4.9120 & 0.5491 & ---\\
Historical prior & 24.8760 & 0.1085 & ---\\
Ordinary pairwise & 30.8044 & 0.5831 & ---\\
Online random forest & 28.8742 & 0.3856 & ---\\
Random forest + residual & 31.2713 & 0.6869 & +2.40\\
\midrule
Online HGB (supplementary) & 29.5501 & 0.3142 & ---\\
HGB + residual (supplementary) & 31.3113 & 1.2779 & +1.76\\
\bottomrule
\end{tabular}
\end{table}

\section{Held-out-source protocol and supporting comparisons}\label{app:heldout}
\paragraph{Cohort and information boundary.}
The 21 confirmation tasks are cancer drug/chemical-response knockout screens selected from the existing benchmark using design metadata and source identity before accessing their outcomes. The retained library-size range is 18,000--22,000. All six recorded sources fall outside the 170-source union covering historical representation training and previous development, confirmation and official-test cohorts. Known aliases and DOI/title matches were checked, together with later usage manifests. Source separation refers to this recorded provenance within the benchmark. All 21 selected tasks were retained, with source counts 2, 1, 1, 11, 3 and 3; correlated task variants contribute through their source average.

The original 65D features, candidate universe, base implementations, residual settings, cached-offset rule and reporting endpoints were fixed before target projection. No confirmation outcomes were used to select these settings or rebuild the representation. The fixed universe covers 97.49--100\% of the confirmation libraries and 99.18--100\% of their positives. FH is defined for every task. The five reported policies share the uniform initial batch within each of three acquisition seeds, then observe only their own selected feedback. These comparisons comprise $21\times5\times3=315$ trajectories, each with 1,000 requests.

\begin{table}[t]
\centering
\caption{Tree-ensemble adaptation and the pairwise control on the held-out-source cohort: 21 tasks, six previously unused recorded sources, three acquisition seeds, and 1,000 requests. Reachable FH and seed SD use the same source-macro aggregation as Table~\ref{tab:heldout}. Gains are relative to each policy's matched base. All 21 tasks have defined reachable FH.}
\label{tab:heldout_all}
\small
\setlength{\tabcolsep}{8pt}
\begin{tabular}{lrr}
\toprule
Policy & Reachable FH (\%) $\pm$ seed SD & Gain (pp) \\
\midrule
Random forest & $17.1853 \pm 0.4700$ & --- \\
Random forest + residual & $19.9836 \pm 1.8659$ & +2.80 \\
\addlinespace[2pt]
HGB & $17.9108 \pm 1.6121$ & --- \\
HGB + residual & $18.6596 \pm 1.8026$ & +0.75 \\
\addlinespace[2pt]
Ordinary pairwise & $20.1659 \pm 3.4608$ & --- \\
\bottomrule
\end{tabular}
\end{table}

\paragraph{Aggregation and reported contrasts.}
Seeds are averaged within task, tasks within recorded source, and sources equally. Seed SD describes acquisition variability. Paired-source bootstrap intervals use 10,000 shared PCG64(20260926) draws and are descriptive 95\% intervals without multiplicity adjustment. The RF own-base gain is +2.80 points [0.95, 5.30]; HGB is +0.75 [$-1.09$, 3.27]. RF improves all six sources, while HGB has three gains and three declines. Source-omission ranges are +1.65 to +3.32 for RF and $-0.44$ to +1.34 for HGB.

Ordinary pairwise fitting reaches 20.17\% FH. The tree-ensemble contrasts therefore establish gains over each matched base; the comparison does not establish a best-performing policy across scorers.

\begin{figure}[htbp]
\centering
\includegraphics[width=\linewidth]{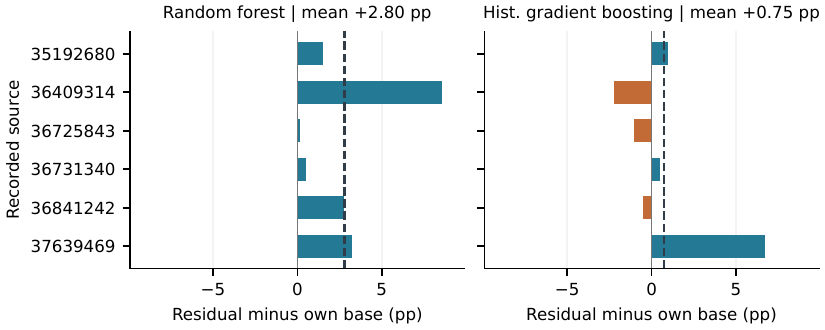}
\caption{Residual-minus-own-base FH differences on all six held-out sources. Each source averages seeds within task and then tasks; dashed lines indicate equally weighted source means. All source effects, including negative HGB effects, are shown.}
\label{fig:heldout_sources}
\end{figure}

\end{document}